\documentclass{article} 
\usepackage{iclr2024_workshop,times}

\usepackage{amsmath,amsfonts,bm}

\def\eqref#1{equation~\ref{#1}}

\def\1{\bm{1}}

\def\ra{{\textnormal{a}}}

\def\rx{{\textnormal{x}}}

\def\rva{{\mathbf{a}}}

\def\erva{{\textnormal{a}}}

\def\ervx{{\textnormal{x}}}

\def\rmA{{\mathbf{A}}}

\def\vmu{{\bm{\mu}}}
\def\vtheta{{\bm{\theta}}}
\def\va{{\bm{a}}}

\def\ve{{\bm{e}}}

\def\vx{{\bm{x}}}

\def\eva{{a}}

\def\mA{{\bm{A}}}

\def\mH{{\bm{H}}}
\def\mI{{\bm{I}}}
\def\mJ{{\bm{J}}}

\def\mX{{\bm{X}}}

\def\mSigma{{\bm{\Sigma}}}

\DeclareMathAlphabet{\mathsfit}{\encodingdefault}{\sfdefault}{m}{sl}
\SetMathAlphabet{\mathsfit}{bold}{\encodingdefault}{\sfdefault}{bx}{n}
\newcommand{\tens}[1]{\bm{\mathsfit{#1}}}
\def\tA{{\tens{A}}}

\def\tX{{\tens{X}}}

\def\gG{{\mathcal{G}}}

\def\sA{{\mathbb{A}}}
\def\sB{{\mathbb{B}}}

\def\sS{{\mathbb{S}}}

\def\emA{{A}}

\newcommand{\etens}[1]{\mathsfit{#1}}

\def\etA{{\etens{A}}}

\newcommand{\E}{\mathbb{E}}

\newcommand{\R}{\mathbb{R}}

\newcommand{\KL}{D_{\mathrm{KL}}}
\newcommand{\Var}{\mathrm{Var}}

\newcommand{\Cov}{\mathrm{Cov}}

\newcommand{\normltwo}{L^2}
\newcommand{\normlp}{L^p}

\newcommand{\parents}{Pa} 

\usepackage{hyperref}
\usepackage{url}
\usepackage{comment}
\usepackage{optidef}
\usepackage{graphicx}
\usepackage{float}

\title{An Adaptive Hydropower Management Approach for Downstream Ecosystem Preservation}

\author{C. Coelho \textsuperscript{1}, M. Jin \textsuperscript{2}, M. Fernanda P. Costa \textsuperscript{1}, L.L. Ferrás \textsuperscript{1,3} \\
\textsuperscript{1}Centre of Mathematics (CMAT), University of Minho \\
\textsuperscript{2}Virginia Tech, Blacksburg, VA 24061 \\
\textsuperscript{3}Department of Mechanical Engineering (Section of Mathematics), FEUP - University of Porto\\
\texttt{cmartins@cmat.uminho.pt, jinming@vt.edu, mfc@math.uminho.pt, lferras@fe.up.pt}
}

\iclrfinalcopy 
\begin{document}

\maketitle

\begin{abstract}
Hydropower plants play a pivotal role in advancing clean and sustainable energy production, contributing significantly to the global transition towards renewable energy sources. However, hydropower plants are currently perceived both positively as sources of renewable energy and negatively as disruptors of ecosystems. In this work, we highlight the overlooked potential of using hydropower plant as protectors of ecosystems by using adaptive ecological discharges. To advocate for this perspective, we propose using a neural network to predict the minimum ecological discharge value at each desired time. Additionally, we present a novel framework that seamlessly integrates it into hydropower management software, taking advantage of the well-established approach of using traditional constrained optimisation algorithms. This novel approach not only protects the ecosystems from climate change but also contributes to potentially increase the electricity production.
\end{abstract}

\section{Introduction}

The link between fossil fuel-based energy production and global warming has spurred intensive exploration into alternative, cleaner energy sources.
Hydropower plants play a pivotal role in advancing clean and sustainable energy production, contributing significantly to the global transition towards renewable energy sources. By harnessing the kinetic energy of flowing water, hydropower offers a reliable and consistent electricity source devoid of greenhouse gas emissions or other pollutants. Its renewable nature, combined with the capacity for large-scale power generation, makes hydropower an important clean energy production strategy.  Furthermore, hydropower plants play a critical role in addressing the intermittent nature of other renewable sources such as solar and wind, improving the stability of the grid, and providing energy storage through reservoirs \citep{chala2019trends}. Compared to fossil fuel alternatives, hydropower has a relatively low environmental footprint, contributing to a decrease in air pollution and efforts to combat climate change. As nations strive to achieve ambitious carbon reduction targets, hydropower stands out as a versatile and environmentally friendly solution, playing a crucial role in building a sustainable energy future \citep{berga2016role}.
However, the construction of hydropower plants within ecosystems demands meticulous consideration of normal flow dynamics to mitigate potential environmental impacts. The ecosystem of a river is a complex web of interconnected relationships between various organisms, water quality parameters, and physical features. The normal flow of a river plays a crucial role in the maintenance of biodiversity and the support of aquatic habitats. When planning and implementing a hydropower project, it is imperative to maintain normal flow patterns to minimise disruptions to the ecosystem. Alteration of flow can affect fish migration, sediment transport, and nutrient distribution, potentially leading to habitat degradation and loss of biodiversity \citep{kuriqi2021ecological}. By preserving normal flow, run-of-river hydropower projects can strike a balance between clean energy generation and environmental sustainability, thus promoting the coexistence of renewable energy development and ecosystem health. 

To mitigate the impact to the ecosystem of hydropower plants, governments mandate minimum discharges, designated by ecological discharges, to promote normal river flow. 
The fixed levels of ecological discharges for maintaining normal river flow are typically established through a combination of scientific research, environmental impact assessments, and regulatory frameworks. Although these levels are intended to protect against detrimental ecological impacts, one challenge is that they may not readily adapt to climate change dynamics. Climate change can lead to alterations in precipitation patterns, temperature regimes, and hydrological cycles, potentially affecting river ecosystems. As climate-related shifts occur, the fixed discharge levels may become less representative of the evolving environmental conditions \citep{renofalt2010effects}. This highlights the need for a dynamic and adaptive regulatory approach that can accommodate the uncertainties associated with climate change, ensuring that environmental discharge standards remain effective in preserving river ecosystems amidst a changing climate.

Non-compliance with environmental regulations can result in fines imposed on hydropower management companies. Consequently, and due to the difficulty of the problem, these companies turn to software tools to optimise water management, enabling them to make optimal decisions over various time periods. This management problem can be formulated as a constrained optimisation problem, with ecological discharges being one of the many constraints to be satisfied \citep{grygier1985algorithms,helseth2020nonconvex}. 

Due to the absence of monetary incentives for allocating water beyond minimum discharges, companies and their management software ultimately strictly adhere to allocating the mandate amount of water. However, the static nature of the imposed ecological discharges may be ill-suited to the dynamic needs of ecosystems, aggravated by factors such as climate change. 
Guidelines may not be adjusted to accommodate these changing needs, leading to a potential insufficiency of water to promote the well-being of habitats caused by periods of drought or when atmospheric conditions are more favourable to water evaporation. Conversely, it may promote unnecessary discharges, especially during wet periods, that could be used for energy generation or storage. The latter underscores how adaptive minimum ecological discharges could benefit hydropower management companies.

Moreover, hydropower plants are currently perceived both positively as sources of renewable energy and negatively as disruptors of ecosystems. In this paper, our primary objective is to underscore the overlooked potential of using hydropower plants as protectors of ecosystems by strategically managing water storage and intelligent discharge. By maintaining the delicate balance of ecosystems during extreme conditions, such as dry periods, hydropower plants can serve as a defence mechanism against the impacts of climate change. To advocate for this perspective, we propose an innovative approach: an adaptive minimum ecological discharge predictor leveraging a neural network (NN). In addition, we present a novel framework that seamlessly integrates it into hydropower management software. Our proposed framework takes advantage of the well-established approach of using traditional constrained optimisation algorithms and uses the adaptive minimum ecological discharge predictor to dictate the minimum discharge value at each solving time.

\section{Proposed Method}

Owing to the problem's high-dimensionality, incorporating numerous variables (energy demand, water availability, turbine efficiency, etc.) and constraints (maximum reservoir capacity, ecological discharges, etc.). This complex problem is typically formulated as the constrained optimisation problem \eqref{eq:p1}, aiming at maximising an objective function $l(\boldsymbol{\theta})$, typically defined by the amount of electricity generated, equating to the company's profit, while satisfying some constraints \citep{fengMultiobjectiveOperationOptimization2017,grygier1985algorithms}. 
Here, we focus on the constraints related to meeting electricity demand and minimum water guidelines for irrigation and ecosystem preservation:

\begin{maxi}|s|[2]
    {\boldsymbol{\theta} \in \mathbb{R}^{n_{\boldsymbol{\theta}}}}{l(\boldsymbol{\theta}),}
    {\label{eq:p1}}
    {}
    \addConstraint{P(\boldsymbol{\theta})}{\geq P_{demand}}
    \addConstraint{Q_{river}(\boldsymbol{\theta})}{\geq Q^{\min}_{river}} 
    \addConstraint{Q_{irrigation}(\boldsymbol{\theta})}{\geq Q^{\min}_{irrigation}},
\end{maxi}

\noindent where $\boldsymbol{\theta} \in \mathbb{R}^{n_{\boldsymbol{\theta}}}$ are the parameters to optimise, $l$ is the objective function, $P$ is the electricity generated, $P_{demand}$ the electricity demand, $Q_{river}$ the amount of water allocated to ecological discharge, $Q^{\min}_{river}$ the minimum water guideline for ecological discharge, $Q_{irrigation}$ the amount of water allocated for irrigation and $Q^{\min}_{irrigation}$ the minimum water guideline for irrigation. 

In this paper, we propose $Q^{\min}_{river}$ to be given by the output of a NN with parameters $\phi$, $Q^{\min}_{river}=NN(\phi)$. This NN predicts the necessary water quantity required to prevent stress on the ecosystem, taking into account anticipated atmospheric conditions, past discharge data and river basin level history. The effectiveness of the predictions relies on the choice of the loss function used to train the model. As an initial guideline, we recommend deviating from the usual prediction versus ground-truth error, given that past discharges may not represent an optimal reference.
Subsequently, employing a traditional optimisation algorithm, \eqref{eq:p1} is solved to obtain the optimal management solution. This optimisation process incorporates the adaptive constraint defined by the minimum ecological discharge provided by the NN, Figure \ref{fig:scheme} in appendix \ref{app:scheme}.

Nevertheless, NNs are considered \emph{black-box} models, making it challenging, and at times impossible, to discern the relationships learned from data. This inherent opacity poses a potential problem, as we cannot risk a situation where a NN provides an exceptionally low minimum ecological discharge value that might stress the ecosystem, simply because abundant rain is anticipated in the near future. Such outcomes could result in fines for management companies, discouraging the adoption of this adaptive strategy. 
To mitigate this potential issue, we propose the explicit integration of expert knowledge constraints into the NN \citep{DBLP:conf/iclr/CoelhoCF23}. The idea is to incorporate inequality constraints on the minimum ecological discharge value, ensuring that scenarios of undue stress on the ecosystem are avoided. We suggest determining the values of these inequality constraints based on the current river basin level and a fixed, invariant to the season, recommended river basin level provided by environmental authorities.

While initially appearing as potentially detrimental to management companies, a closer examination reveals otherwise. Relying on fixed minimum ecological discharges can lead to unnecessary disposal of water that could otherwise be used for electricity generation, particularly under conditions of minimal water evaporation or the presence of rainfall. These factors contribute to reducing the necessity for ecological discharges, presenting an opportunity for profit increase.
During dry summer periods when fixed ecological discharges might prove insufficient for maintaining downstream ecosystem health, our proposed framework should dynamically adjust the constraint, allocating less water for electricity generation, potentially decreasing profits, and possibly jeopardising the ability to meet electricity demand. However, during dry seasons characterised by intense sunlight, we recommend combining our proposed method with solar panels. Despite increased water allocation for ecological discharges, electricity demand and company profits can be sustained through the use of solar panels for power generation. This integrated approach ensures adaptability and resilience for hydropower management companies under varying climatic conditions. To incentivise companies to adopt this approach, we encourage governments to provide incentives, such as financial support for the expenses associated with implementing solar panels \citep{solar}.

\section{Conclusion}
In this preliminary study, we underscore the challenges associated with fixed constraints pertaining to minimum ecological discharges in hydropower management software. The current approach of employing static guidelines, incapable of adapting promptly to atmospheric conditions, poses a threat to ecosystems, especially during unexpected dry seasons. Additionally, enforcing fixed minimum ecological discharges compels management companies to adhere rigidly to these standards, leading to potential misalignment during wet periods when the allocated water might not be necessary. In such cases, to avoid fines, water is wasted instead of being used for electricity generation or stored for future use. This inflexibility highlights the need for a more adaptive and dynamic approach to ecological discharge management in hydropower systems. 
To promote this idea, we propose a framework that still uses a traditional constrained optimisation algorithm to solve the management problem that receives the minimum ecological discharge value from a NN that is able to, using atmospheric conditions and previous ecological discharges, predict the amount of discharge needed for promoting a healthy ecosystem. 

Our proposal encourages a paradigm shift, portraying hydropower plants as substantial batteries capable of generating clean energy while concurrently promoting ecosystem well-being. This perspective offers a pathway to mitigate the impact of climate change on the environment. 

\bibliography{iclr2024_conference}
\bibliographystyle{iclr2024_conference}

\newpage
\appendix
\section{Scheme of the proposed method} \label{app:scheme}

\begin{figure}[h]
    \centering
    \includegraphics[scale=0.50]{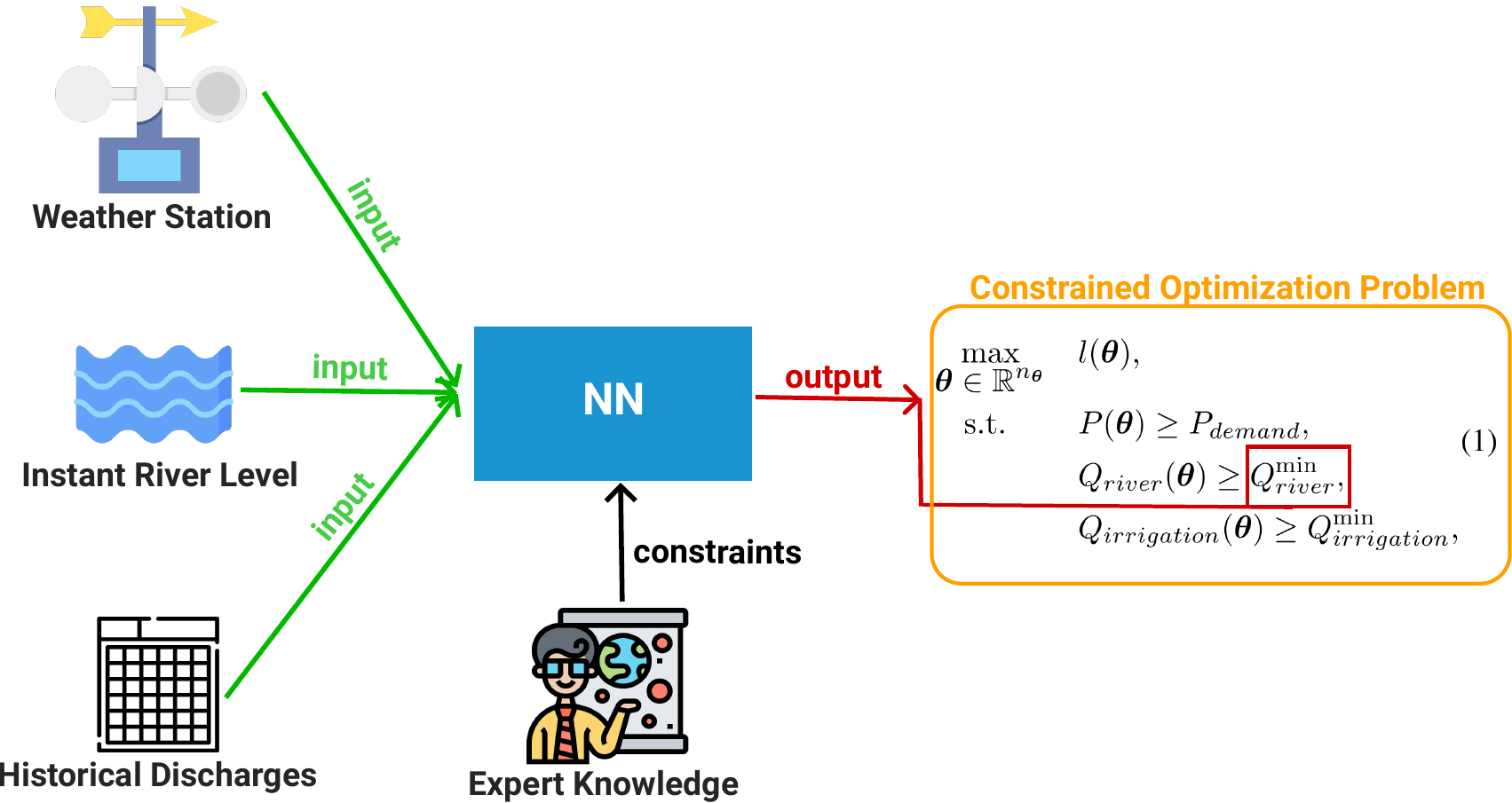}
    \caption{Graphical representation of the proposed method.}
    \label{fig:scheme}
\end{figure}

\end{document}